\documentclass[letterpaper]{article} % DO NOT CHANGE THIS
\usepackage{aaai2027}  % DO NOT CHANGE THIS
\usepackage[hyphens]{url}  % DO NOT CHANGE THIS
\usepackage{graphicx} % DO NOT CHANGE THIS
\usepackage{natbib}  % DO NOT CHANGE THIS AND DO NOT ADD ANY OPTIONS TO IT
\usepackage{caption} % DO NOT CHANGE THIS AND DO NOT ADD ANY OPTIONS TO IT
\usepackage{algorithm}
\usepackage{algorithmic}

\usepackage{newfloat}
\usepackage{listings}
\DeclareCaptionStyle{ruled}{labelfont=normalfont,labelsep=colon,strut=off} % DO NOT CHANGE THIS
\floatstyle{ruled}
\newfloat{listing}{tb}{lst}{}
\floatname{listing}{Listing}

\usepackage{booktabs}
\usepackage{amsmath}
\usepackage{amssymb}

\title{DINS-IO: Learned Inertial Odometry via Differentiable INS Consistency}
\author{
    Hao Qiao\textsuperscript{\rm 1},
    Yan Wang\textsuperscript{\rm 2}\corresponding,
    Jian Kuang\textsuperscript{\rm 2},
    Xiaoji Niu\textsuperscript{\rm 2,\rm 3}
}
\affiliations{
    \textsuperscript{\rm 1}School of Geodesy and Geomatics, Wuhan University\\
    \textsuperscript{\rm 2}GNSS Research Center, Wuhan University\\
    \textsuperscript{\rm 3}Hubei Luojia Laboratory\\
    \{hmqiao, wystephen, kuang, xjniu\}@whu.edu.cn
}

\begin{document}
\nocopyright
\maketitle

\begin{abstract}
The training of learned inertial odometry depends on dense, high-precision position ground truth from motion capture, visual-inertial odometry or SLAM, which is costly and hard to acquire at scale. We propose \textbf{DINS-IO}, which learns inertial odometry directly from raw IMU streams \emph{without position labels}. Our key insight is that the strapdown INS velocity recursion is a strong, fully differentiable consistency prior: the predicted velocity, rotated into the navigation frame, must agree with the integrated specific force up to an unknown initial velocity and a constant accelerometer bias. We cast this constraint as a sliding-window least-squares problem with a globally shared bias, solve it in closed form, and use the solver residual as a self-supervised loss whose gradient flows back to the network through the analytic solution. To supply this per-sample constraint, we design a high-frequency network that emits dense body-frame velocity at the IMU rate. Since the self-supervised network learns consistent motion but its velocity is not yet metrically calibrated, we calibrate it to true metric velocity from a few labeled trajectories by directly supervising the predicted body-frame velocity and adapting only low-rank (LoRA) patches. On standard benchmarks, DINS-IO pretrained self-supervised and fine-tuned with a small fraction of labels matches or surpasses fully supervised baselines.
\end{abstract}


\section{Introduction}

Inertial odometry estimates the motion of a platform from a low-cost inertial measurement unit (IMU). Because IMUs are small, cheap, power-efficient, and independent of any external infrastructure, they are ubiquitous in smartphones, wearables, robots, and drones. Classical strapdown inertial navigation integrates the gyroscope and accelerometer signals through the kinematic equations of motion, but the double integration of sensor noise and bias causes the position error to grow cubically with time, making pure inertial integration unusable beyond a few seconds for low-cost sensors.

Learned inertial odometry (LIO) sidesteps this brittleness by training a neural network to regress motion (velocity or displacement) from short windows of IMU data, implicitly exploiting the statistical regularities of human and platform motion. This data-driven approach has dramatically improved drift characteristics and now underpins many practical pedestrian and platform tracking systems. Yet essentially all high-performing LIO models are trained under \emph{full supervision}: they require dense, high-precision position or velocity ground truth, typically obtained from a motion-capture rig, a visual-inertial system, or offline SLAM. This dependence is the central bottleneck of the paradigm:

\begin{itemize}
    \item \textbf{Cost and scale.} High-precision position ground truth needs specialized hardware or calibrated reference systems, so labeled inertial datasets stay small relative to the vast \emph{unlabeled} IMU data any device produces continuously.
    \item \textbf{Limited coverage and domain bias.} Reference systems work only under constrained conditions (motion-capture studios, textured scenes for visual-inertial odometry, places where SLAM converges), so labeled trajectories are biased toward a narrow slice of environments and motions, and supervised models degrade in the in-the-wild conditions that matter most.
    \item \textbf{No physical grounding.} Trained only to match labels, the regressor fits the labeled dataset's motion statistics rather than the underlying strapdown kinematics, giving no consistency guarantee and weak generalization across users, carrying modes, and platforms.
\end{itemize}

We argue that the IMU signal already contains a powerful supervisory signal that has been overlooked: the laws of inertial navigation themselves. Although double integration is too fragile to \emph{produce} a trajectory, the underlying strapdown velocity recursion still imposes a tight, exact relationship between the measured specific force, the platform attitude, and the platform velocity. Any velocity sequence the network predicts must be \emph{physically consistent} with the measured specific force, up to two small unknowns: the initial velocity of the window and a constant accelerometer bias. This consistency holds regardless of whether a position label is available.

\begin{figure}[t]
\centering
\includegraphics[width=\columnwidth]{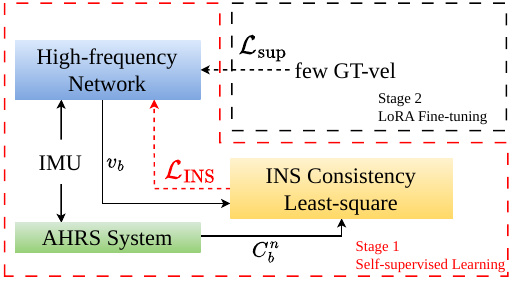}
\caption{Overview of DINS-IO. \textbf{Stage~1} (no labels): the network predicts dense body-frame velocity that must satisfy the strapdown INS recursion via a differentiable least-squares (LS) solve, whose residual is the loss. \textbf{Stage~2}: a few labeled trajectories supervise the predicted body-frame velocity directly, and only LoRA adapters are trained to calibrate it to metric velocity.}
\label{fig:overview}
\end{figure}

Building on this observation we propose \textbf{DINS-IO}, a self-supervised-then-label-efficient framework for inertial odometry (Figure~\ref{fig:overview}). Our contributions are:

\begin{itemize}
    \item We propose a differentiable INS-consistency loss that casts the strapdown velocity recursion as a sliding-window least-squares problem and uses its closed-form residual as a self-supervised objective, training the network from raw IMU data without any position or velocity label.
    \item We calibrate this self-supervised prediction to metric scale from only a few labeled trajectories: a high-frequency network emits dense body-frame velocity at the IMU rate, providing the per-sample signal required by the consistency loss and LoRA adapters supervise this velocity directly, reaching metric accuracy without overwriting the physically grounded features from pretraining.
\end{itemize}

Together these components decouple model quality from the availability of expensive position ground truth.

\section{Related Work}

\paragraph{Learned inertial odometry.}
Data-driven inertial odometry replaces fragile double integration with a network that regresses motion from short IMU windows. IONet~\citep{chen2018ionet} pioneered this idea by learning polar displacement increments with a recurrent network, and RoNIN~\citep{herath2020ronin} established strong velocity-regression baselines together with a large benchmark. TLIO~\citep{liu2020tlio} tightly couples a learned displacement estimator with a stochastic filter, while IDOL~\citep{sun2021idol} jointly learns orientation and position to reduce reliance on external attitude. More recent work improves the regression backbone: CTIN~\citep{rao2022ctin} introduces a contextual transformer, M2EIT~\citep{li2025m2eit} uses a multi-domain mixture of experts for robustness across motion modes, and lightweight designs such as LLIO~\citep{wang2022llio}, IMUNet~\citep{zeinali2024imunet}, and ResMixer~\citep{lai2024resmixer} target efficient on-device deployment. AirIO~\citep{qiu2025airio} enhances IMU feature observability and predicts velocity in the body frame, an output convention we share, while TartanIMU~\citep{zhao2025tartan} trains a light foundation model for inertial positioning and adapts it to new robots, pointing toward a pretrain-then-adapt paradigm. Despite their architectural diversity, all of these methods still rely on \emph{supervised} position or velocity ground truth---obtained from motion capture, visual-inertial systems, or SLAM---for both training and adaptation. This dependence caps the data they can exploit and is exactly the bottleneck DINS-IO removes: we pretrain with the differentiable INS-consistency residual instead of any position label, and calibrate the prediction to metric velocity from only a small labeled subset.

\paragraph{Self-supervised learning for inertial sensing.}
To escape the label bottleneck, a growing body of work pretrains on unlabeled inertial data with self-supervised pretext tasks. In human-activity recognition, masking-and-reconstruction (e.g.\ LIMU-BERT~\citep{xu2021limu}) and contrastive objectives (contrastive predictive coding~\citep{haresamudram2021contrastive}, and cross-dataset variants such as CrossHAR~\citep{hong2024crosshar} and ContrastSense~\citep{dai2024contrastsense}) learn transferable representations that are then fine-tuned on a small labeled set; several of these inject the physics of the sensing process into the augmentations to bridge device and user gaps. Closer to odometry, RIO~\citep{cao2022rio} uses rotation-equivariance as a self-supervision signal to improve generalization without labels, while physics-informed formulations such as SSPINNpose~\citep{gambietz2025sspinnpose} enforce dynamics and temporal-consistency losses, and masking-based denoisers~\citep{yang2025r} reconstruct redundant IMU streams. These methods, however, learn either a generic representation or an auxiliary regularizer: the pretext task is not the navigation geometry itself, so a metric motion estimate still has to be learned from labels downstream. In contrast, our pretext task \emph{is} the strapdown velocity recursion, so self-supervision directly yields a body-frame velocity---consistent in direction and temporal profile, though not yet metrically calibrated---rather than a representation that must be decoded by a supervised head.

\paragraph{Parameter-efficient fine-tuning.}
Adapting a large pretrained model to a downstream task by updating only a small number of parameters has become standard practice. Low-Rank Adaptation (LoRA)~\citep{hu2022lora} freezes the pretrained weights and learns a low-rank update $\Delta W = BA$ with a zero-initialized $B$, so fine-tuning starts as an identity map and touches a tiny fraction of parameters; variants such as QLoRA~\citep{dettmers2023qlora} and DoRA~\citep{liu2024dora} extend it further, and adapter-style tuning has proven effective well beyond language, including vision and inertial odometry models~\citep{zhao2025resilient}. We adopt LoRA for a different reason than model size: our labeled data is scarce, so restricting the update to low-rank patches lets the few labeled trajectories calibrate the prediction to metric velocity without overwriting the physically grounded features learned during self-supervised pretraining.

\section{Preliminaries}

\subsection{Strapdown INS Velocity Recursion}

Let $f^b[k]\in\mathbb{R}^3$ be the specific force measured by the accelerometer at time index $k$ (body frame, $\mathrm{m/s^2}$), $R[k]\in SO(3)$ the rotation from body to navigation frame (obtained from an attitude/AHRS estimate), $b_a\in\mathbb{R}^3$ a constant accelerometer bias, and $g^n$ gravity expressed in the navigation frame. The continuous-time navigation-frame velocity obeys
\begin{equation}
\dot v^n = R\,(f^b - b_a) + g^n
\end{equation}
Discretizing with step $\Delta t$ and integrating from the start of a window to index $k$ gives
\begin{equation}
\begin{aligned}
v^n[k] = v^n_0 &+ \sum_{i=0}^{k-1} R[i]\,f^b[i]\,\Delta t \\
&- \Big(\sum_{i=0}^{k-1} R[i]\,\Delta t\Big) b_a + g^n k \Delta t
\end{aligned}
\end{equation}
Introducing the three cumulative quantities
\begin{align}
S_f[k] &= \textstyle\sum_{i<k} R[i]\,f^b[i]\,\Delta t \\
S_R[k] &= \textstyle\sum_{i<k} R[i]\,\Delta t \\
g_{\mathrm{corr}}[k] &= g^n\,k\,\Delta t
\end{align}
the recursion collapses to the compact linear form
\begin{equation}
\,v^n[k] = v^n_0 + S_f[k] - S_R[k]\,b_a + g_{\mathrm{corr}}[k]\,
\label{eq:ins}
\end{equation}
Equation~\eqref{eq:ins} is exact and \emph{linear} in the two unknowns $(v^n_0, b_a)$; everything else is computed directly from the measured specific force, the attitude, and known gravity.

\subsection{Problem Setup}

The network consumes a stream of IMU data (and the associated attitude) and outputs, for every sample $k$, a body-frame velocity $v^b_{\mathrm{pred}}[k]\in\mathbb{R}^3$. Downstream localization rotates $v^b_{\mathrm{pred}}$ into the navigation frame and integrates it to obtain a trajectory; hence both the \emph{direction} and the \emph{absolute magnitude} of the predicted velocity must be accurate. Our goal is to learn this mapping primarily from unlabeled IMU data, using position ground truth for only a small fraction of trajectories.

\section{Method}

\begin{figure*}[t]
\centering
\includegraphics[width=\textwidth]{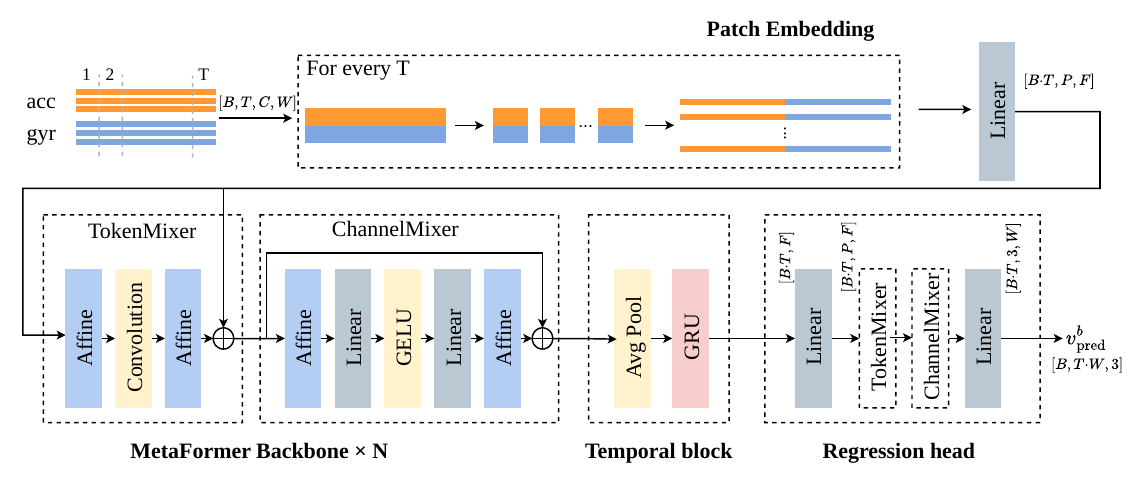}
\caption{Network architecture of DINS-IO. A three-stage ``per-window feature extraction $\rightarrow$ temporal modeling $\rightarrow$ high-frequency regression'' design maps windowed IMU data to a dense body-frame velocity sequence at every IMU sample.}
\label{fig:network}
\end{figure*}

\subsection{Overview}

DINS-IO has two stages that share one network backbone and one least-squares (LS) machinery (Figure~\ref{fig:overview}). In \textbf{Stage 1} the network is pretrained purely self-supervised: predicted velocities are required to satisfy the INS recursion of Eq.~\eqref{eq:ins} through a differentiable LS solve, and the residual is the loss. In \textbf{Stage 2} a small amount of position-labeled data metrically calibrates the predicted body-frame velocity by directly supervising it, while only low-rank adapters are updated. We first describe the network that makes a per-sample loss possible, then the self-supervised loss, why it leaves the velocity metrically underdetermined, and finally the label-efficient calibration stage.

\subsection{High-Frequency Inertial Velocity Network}

The INS observation in Eq.~\eqref{eq:ins} is defined \emph{per IMU sample}: $S_f$ and $S_R$ accumulate the per-sample specific force and attitude. To participate in this constraint, the network must therefore output a body-frame velocity at every IMU sample, i.e.\ a dense high-frequency sequence, rather than a single per-window displacement as in typical LIO regressors. We adopt a three-stage ``per-window feature extraction $\rightarrow$ temporal modeling $\rightarrow$ high-frequency regression'' design (Figure~\ref{fig:network}).

\paragraph{Input and patch embedding.}
The input is organized as $[B,T,C,W]$: $B$ sequences, $T$ one-second time steps, $C$ inertial channels, and $W$ samples per second (e.g.\ $W{=}100$ at $100$\,Hz). Each one-second window is split into patches and linearly projected to a feature of dimension $F$, yielding $[B{\cdot}T, P, F]$ with $P$ patches per window.

\paragraph{MetaFormer backbone.}
A stack of MetaFormer blocks~\cite{yu2022metaformer} models intra-window structure with the residual pattern $x \leftarrow x + \mathrm{TokenMixer}(x)$ and $x \leftarrow x + \mathrm{ChannelMixer}(x)$. The token mixer exchanges information across patches; the channel mixer is an expand-contract MLP. Per-layer initialization scaling $1/(i{+}1)$ stabilizes training of the deeper layers.

\paragraph{Temporal block.}
A reduction over the patch dimension produces one feature vector per second, $[B,T,F]$, which a recurrent (GRU) block processes along $T$ to capture cross-second temporal dependencies. The block supports both a batched forward for training and a single-step forward for streaming inference, so that training and online deployment share weights and produce identical results.

\paragraph{High-frequency regression head.}
The crucial component for our loss is the regression head. From each per-second temporal feature, a \emph{time-generator} linear layer re-expands the single vector into $P$ patch tokens, and a head built on a single MetaFormer block maps these back to a full-rate window of length $W$. The head therefore turns one temporal feature into a dense $W$-sample body-frame velocity prediction. Stacking over $T$ gives the dense prediction $v^b_{\mathrm{pred}}\in\mathbb{R}^{B\times (T W)\times 3}$ consumed by the consistency loss. This ``short-feature $\rightarrow$ long-output'' expansion is what lets a compact temporal model still emit the high-rate velocity the differentiable INS observation requires.

\subsection{Differentiable INS Consistency for Self-Supervision}

\paragraph{Linear observation.}
Given the dense prediction, we rotate it into the navigation frame to form an \emph{observation}
\begin{equation}
v^n_{\mathrm{obs}}[k] = R[k]\,v^b_{\mathrm{pred}}[k]
\end{equation}
If the prediction is correct, $v^n_{\mathrm{obs}}[k]$ must satisfy Eq.~\eqref{eq:ins}. Collecting the unknowns into $x=[\,v^n_0;\,b_a\,]\in\mathbb{R}^6$ and moving known terms to one side yields, for each $k$, a linear observation
\begin{equation}
\underbrace{v^n_{\mathrm{obs}}[k] - S_f[k] - g_{\mathrm{corr}}[k]}_{z[k]\in\mathbb{R}^3}
=
\underbrace{[\,I_3\ \mid\ -S_R[k]\,]}_{H[k]\in\mathbb{R}^{3\times 6}}\,x
\label{eq:obs}
\end{equation}

\paragraph{Sliding-window joint LS with a shared bias.}
Integrating a long sequence as a single window lets the cumulative $S_f$ accumulate sensor noise. We therefore split the sequence of length $T$ into $W$ non-overlapping windows of length $\mathit{ws}$, re-accumulating $S_f$ and $S_R$ \emph{locally} inside each window. Crucially, we let every window share \emph{one} global accelerometer bias $b_a$, stacking the unknowns as
\begin{equation}
X = [\,v^n_{0,0};\,v^n_{0,1};\,\dots;\,v^n_{0,W-1};\,b_a\,]\in\mathbb{R}^{3W+3}
\end{equation}
A shared bias reduces the unknown count from $6W$ (independent per-window biases) to $3W{+}3$ and, more importantly, makes $b_a$ jointly constrained by \emph{all} observations rather than letting each per-window bias silently absorb a systematic error. We assemble two groups of rows:
\begin{itemize}
    \item \textbf{Within-window INS rows} ($W\!\cdot\!\mathit{ws}\!\cdot\!3$ rows): Eq.~\eqref{eq:obs} for every sample, with $I_3$ placed in the block of window $w$ and $-S_{R,w}[k]$ in the shared-bias block.
    \item \textbf{Boundary continuity rows} ($(W{-}1)\!\cdot\!3$ rows): adjacent windows must agree at their shared boundary, i.e.\ the end velocity reconstructed in window $w$ equals the start velocity $v^n_{0,w+1}$ of the next, coupling consecutive initial velocities through the shared bias.
\end{itemize}
Stacking these into $H_{\mathrm{full}}$ and $z_{\mathrm{full}}$, the LS solution is the closed form
\begin{equation}
X^\star = \big(H_{\mathrm{full}}^\top H_{\mathrm{full}} + \lambda I\big)^{-1} H_{\mathrm{full}}^\top z_{\mathrm{full}}
\label{eq:solve}
\end{equation}
with a small Tikhonov term $\lambda$ for numerical stability. The self-supervised loss is the residual
\begin{equation}
\ \mathcal{L}_{\mathrm{INS}} = \tfrac{1}{M}\,\big\| H_{\mathrm{full}}\,X^\star - z_{\mathrm{full}} \big\|^2\
\label{eq:loss}
\end{equation}
where $M$ is the number of rows.

\paragraph{Why it is differentiable and label-free.}
$z_{\mathrm{full}}$ contains $v^n_{\mathrm{obs}} = R\,v^b_{\mathrm{pred}}$, a linear function of the network output. The solution $X^\star$ in Eq.~\eqref{eq:solve} is an analytic (and differentiable) function of $z_{\mathrm{full}}$, so $\mathcal{L}_{\mathrm{INS}}$ is differentiable with respect to $v^b_{\mathrm{pred}}$ and its gradient back-propagates to the network. If the prediction exactly satisfies the INS recursion, some $(v^n_0,b_a)$ drives the residual to zero, so $\mathcal{L}_{\mathrm{INS}}=0$. No position or velocity label appears anywhere in Eqs.~\eqref{eq:obs}--\eqref{eq:loss}.

\subsection{Label-Efficient Metric Calibration via LoRA}

\paragraph{Direct body-frame velocity supervision.}
Metric velocity is supplied most directly by velocity labels. For the small labeled subset, the labeled trajectory provides a ground-truth navigation-frame velocity, which we rotate into the body frame to obtain a per-sample target matching the network's output convention,
\begin{equation}
v^b_{\mathrm{gt}}[k] = \big(R[k]\big)^\top v^n_{\mathrm{gt}}[k]
\end{equation}
We then fine-tune with a direct supervised loss on the dense prediction,
\begin{equation}
\mathcal{L}_{\mathrm{sup}} = \frac{1}{N}\sum_{k} \big\| v^b_{\mathrm{pred}}[k] - v^b_{\mathrm{gt}}[k] \big\|^2
\label{eq:sup}
\end{equation}
where $N$ is the number of supervised samples. Because the target carries the true metric magnitude, this loss calibrates $v^b_{\mathrm{pred}}$ to metric velocity directly, exactly what the self-supervised INS residual leaves unobserved.

\paragraph{LoRA fine-tuning.}
Labeled trajectories are scarce, so full fine-tuning would both overfit and risk catastrophically forgetting the self-supervised features. We instead apply Low-Rank Adaptation: a linear layer $y = Wx+b$ is augmented as
\begin{equation}
y = Wx + b + \tfrac{\alpha}{r}\,B A\,x
\end{equation}
with $W,b$ frozen, $A\in\mathbb{R}^{r\times \mathrm{in}}$ Kaiming-initialized, $B\in\mathbb{R}^{\mathrm{out}\times r}$ zero-initialized, and rank $r\ll\min(\mathrm{in},\mathrm{out})$. The zero initialization of $B$ makes the adapted network identical to the pretrained one at the start of fine-tuning, giving a smooth, drift-free hand-off.

We insert adapters into the linear layers of the high-frequency regression head only, and freeze the input layer, the MetaFormer backbone, and the temporal model entirely. This placement follows the division of labor established during pretraining: the backbone and temporal model already encode the direction and temporal profile of motion from abundant unlabeled data, and self-supervision leaves only the metric magnitude undetermined---a low-dimensional, largely multiplicative calibration that the head, which maps temporal features to the dense body-frame velocity, is best positioned to absorb. Restricting adaptation to the head thus keeps the physically grounded representation frozen, further shrinks the trainable set (under $10\%$ of parameters), and guards against overfitting the scarce labels. The supervised velocity loss of Eq.~\eqref{eq:sup} supplies the metric-calibration gradient, so a handful of labeled trajectories suffice to recover metric velocity while the inductive bias learned from unlabeled IMU data stays intact.

\section{Experiments}

We evaluate DINS-IO around three questions: (i) whether the label-free INS-consistency objective alone learns a usable motion estimate; (ii) how many labels the LoRA metric-calibration stage needs to match or surpass fully supervised inertial odometry; and (iii) which design choices are responsible for the result.

\subsection{Experimental Setup}

\paragraph{Datasets.}
We use two pedestrian inertial datasets. \textbf{TLIO}~\citep{liu2020tlio} provides IMU streams at $100$\,Hz together with high-precision position and orientation ground truth from a visual-inertial system, and is a standard benchmark for learned inertial odometry. \textbf{Tango} is our self-collected dataset, recorded on the ASUS Tango phone, whose fisheye global-shutter camera and embedded IMU yield a visual-inertial trajectory that we use as ground truth (with area learning disabled for smoothness). It comprises more than $40$ hours of head-mounted pedestrian data from six subjects across multiple physical devices, spanning varied activities (walking, standing, sitting, and stair climbing) and thus diverse motion patterns and device-specific IMU errors. For each dataset we hold out a disjoint set of complete trajectories for testing, treat the remaining trajectories \emph{without their labels} as the Stage~1 pretraining pool, and draw the small Stage~2 labeled subsets from that same pool.

\paragraph{Metrics.}
We use the following metrics for evaluation. Let $\hat{p}_i$ and $p_i$ denote the estimated and ground-truth positions at sample $i$, and $N$ the number of samples in a trajectory. The Absolute Trajectory Error (ATE) is the root-mean-square position error over the whole trajectory,
\begin{equation}
\mathrm{ATE} = \sqrt{\frac{1}{N}\sum_{i=1}^{N}\big\| p_i - \hat{p}_i\big\|^2}
\label{eq:ate}
\end{equation}
The Relative Trajectory Error (RTE) measures drift over a fixed time interval of $\Delta t$ samples ($60$\,s). Within each window we align the estimate to the ground truth at the window start by the relative orientation $R_i\hat{R}^{T}_i$, where $R_i$ and $\hat{R}_i$ are the ground-truth and estimated orientations, and compare the two displacements,
\begin{equation}
\mathrm{RTE} = \sqrt{\frac{1}{N}\sum_{i=1}^{N}\big\|(p_{i + \Delta t} - p_i) - R_i \hat{R}^{T}_i(\hat{p}_{i + \Delta t} - \hat{p}_i)\big\|^2}
\label{eq:rte}
\end{equation}
For the label-free analysis of Stage~1 we additionally report the velocity-direction error, the per-sample angle between the predicted and ground-truth navigation-frame velocity, evaluated only on samples whose ground-truth speed exceeds $0.15$\,m/s, since the direction is undefined at rest.

\paragraph{Baselines.}
We compare against representative supervised LIO models: the three RoNIN backbones~\citep{herath2020ronin} (ResNet, LSTM, and TCN), the lightweight LLIO~\citep{wang2022llio}, and the body-frame velocity model AirIO~\citep{qiu2025airio}, whose output convention we share. We further include DINS-IO trained under full supervision from scratch, which shares our backbone but omits self-supervised pretraining and thus isolates the contribution of Stage~1. All learned baselines are trained on the same labeled subsets used by our Stage~2; only DINS-IO additionally exploits the unlabeled pretraining pool.

\subsection{Self-Supervision Recovers Velocity Direction}

\begin{figure*}[b]
\centering
\includegraphics[width=0.48\linewidth]{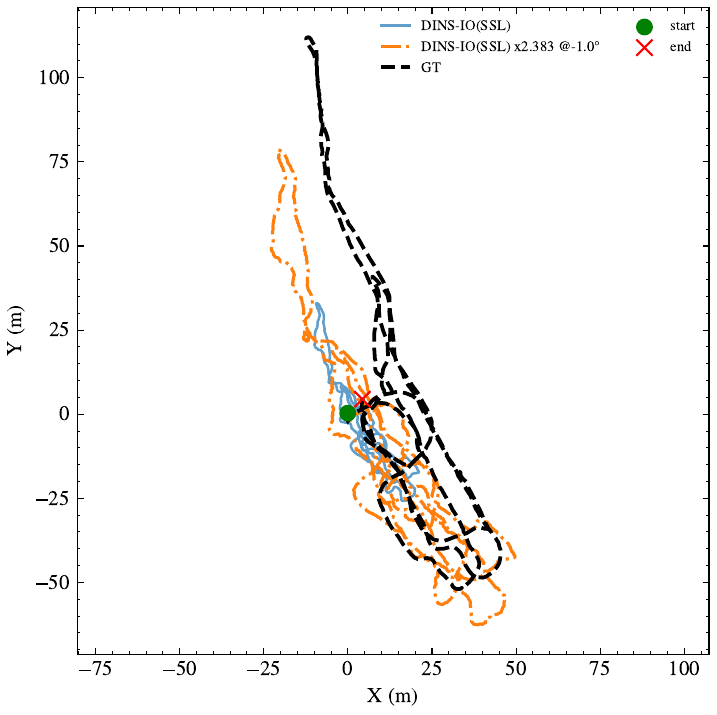}\hfil
\includegraphics[width=0.48\linewidth]{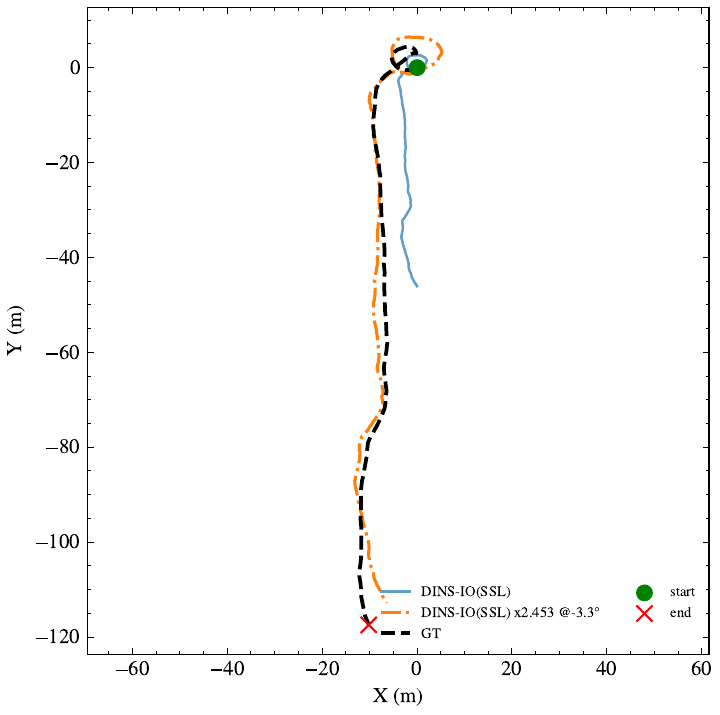}\\
\includegraphics[width=0.48\linewidth]{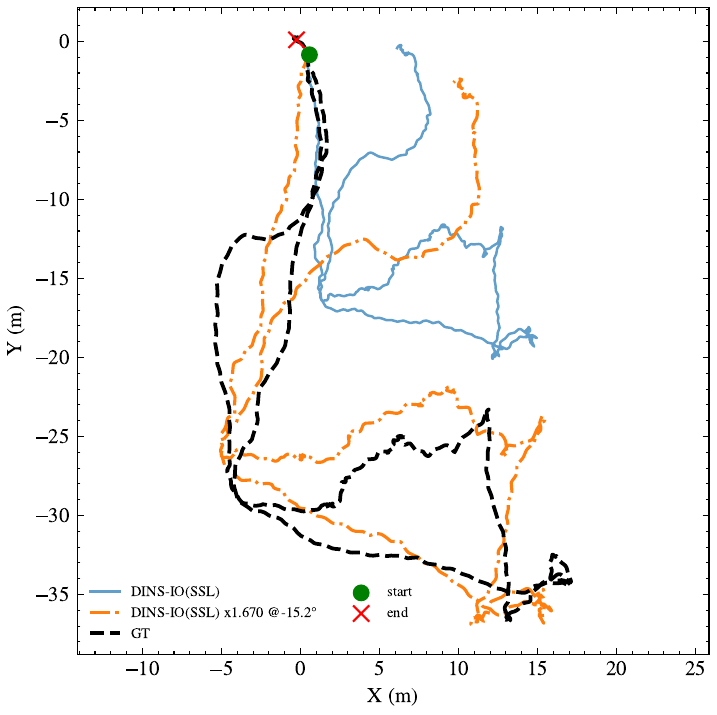}\hfil
\includegraphics[width=0.48\linewidth]{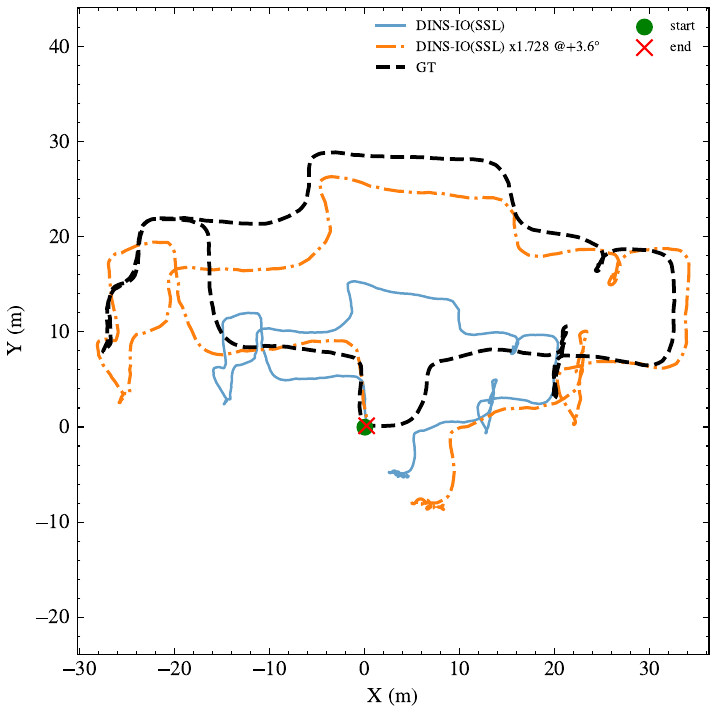}
\caption{Trajectories from integrating the label-free Stage~1 velocity (blue) against ground truth (black dashed), for two Tango (top) and two TLIO (bottom) sequences. The orange curve is the same Stage~1 trajectory after a similarity transform (rotation and scaling only, no translation) aligns it to the ground truth, isolating how well its shape and heading match once the uncalibrated scale is removed. Self-supervision alone recovers the coarse shape and heading of the path.}
\label{fig:ssl_traj}
\end{figure*}

We first ask whether Stage~1, trained without any label, already captures motion. We take the pretrained model \emph{before any fine-tuning} and examine the direction of its predicted velocity, rotated into the navigation frame $v^n_{\mathrm{pred}}=R\,v^b_{\mathrm{pred}}$, against the ground-truth navigation-frame velocity. Across the test set (Table~\ref{tab:dir}) the label-free model attains a median direction error of 14.0\textdegree{} on TLIO and 21.1\textdegree{} on Tango, with more than 82.4\% and 69.1\% of samples within $30$\textdegree{} in two datasets. The direction error stays low throughout a sequence rather than degrading over time, showing that the estimate is stable and not merely accurate on average.
\begin{table}[t]
\centering
\small
\begin{tabular}{ll ccc}
\toprule
Dataset & Method & Mean & Median & $<30$\textdegree \\
        &        & (\textdegree) & (\textdegree) & (\%) \\
\midrule
TLIO  & DINS-IO (SSL, no FT) & 20.3 & 14.0 & 82.4 \\

Tango & DINS-IO (SSL, no FT) & 27.4 & 21.1 & 69.1 \\
\bottomrule
\end{tabular}
\caption{Velocity-direction error of the label-free Stage~1 model, over samples with $\|v^n_{\mathrm{gt}}\|>0.15$\,m/s}
\label{tab:dir}
\end{table}

This directional consistency is visible at the trajectory level. Figure~\ref{fig:ssl_traj} integrates the raw, un-calibrated velocity of the label-free Stage~1 model into a position track. On both datasets the reconstruction reproduces the coarse shape and heading of the ground-truth path---turns occur where they should and the early portion of each sequence overlaps the reference---confirming that self-supervision alone already captures the geometry of motion. Because the velocity is not yet metrically calibrated, however, the integrated scale is off (most clearly on Tango, where the recovered path is compressed relative to the true multi-loop trajectory). This is exactly the residual degree of freedom that INS consistency leaves unobserved and that Stage~2 calibrates from a few labels.

\section{Conclusion}

We presented DINS-IO, which learns inertial odometry from raw IMU data by turning the strapdown INS velocity recursion into a differentiable self-supervised loss, supported by a high-frequency network that emits the dense body-frame velocity the loss requires. Because INS consistency leaves the metric velocity underdetermined under degenerate motion, we calibrate it with a small labeled subset by directly supervising the predicted body-frame velocity and training only low-rank adapters. This decouples model quality from the availability of expensive position ground truth, enabling inertial-odometry models to be trained primarily from abundant unlabeled data.

% References and End of Paper
\bibliography{aaai2027}

\end{document}